\documentclass[lettersize,journal]{IEEEtran}
\usepackage{amsmath,amsfonts}
\usepackage{algorithmic}
\usepackage{algorithm}
\usepackage{array}
\usepackage{textcomp}
\usepackage{stfloats}
\usepackage{url}
\usepackage{verbatim}
\usepackage{graphicx}
\usepackage{cite}
\usepackage{hyperref}
\usepackage{adjustbox}

\usepackage{algorithm}
\usepackage{algorithmic}
\usepackage{booktabs}
\usepackage{amsmath}
\usepackage{graphicx}
\usepackage{subfigure}
\usepackage{amsfonts,bm}
\usepackage{multirow}
\usepackage{color}
\usepackage{braket}
\usepackage{mathtools}
\usepackage{enumerate}
\usepackage{enumitem}
\usepackage{empheq}
\usepackage{calc}
\usepackage{dsfont}
\usepackage{url}
\usepackage[flushmargin]{footmisc}%hang,

\usepackage{placeins}

\let\vec\textbf

\newcommand{\vx}{\vec{X}}

\newcommand{\vy}{\vec{Y}}

\begin{document}

\title{
When are Foundation Models Effective? 
Understanding the Suitability for Pixel-Level Classification Using Multispectral Imagery
}

\author{
Yiqun Xie, Zhihao Wang, Weiye Chen, Zhili Li, Xiaowei Jia, Yanhua Li, \\Ruichen Wang, Kangyang Chai, Ruohan Li, Sergii Skakun
\vspace{-20pt}
\thanks{
Y. Xie, Z. Wang, W. Chen, Z. Li, R. Wang, K. Chai, R. Li and S. Skakun are with University of Maryland.
E-mail:  \{xie, zhwang1, weiyec, lizhili, r526li, ruichenw, kychai, skakun\}@umd.edu.
X. Jia is with University of Pittsburgh.
E-mail:  xiaowei@pitt.edu.
Y. Li is with Worcester Polytechnic Institute.
E-mail:  yli15@wpi.edu.
}
}



\maketitle

\begin{abstract}
Foundation models, i.e., very large deep learning models, have demonstrated impressive performances in various language and vision tasks that are otherwise difficult to reach using smaller-size models. The major success of GPT-type of language models is particularly exciting and raises expectations on the potential of foundation models in other domains including satellite remote sensing.
In this context, great efforts have been made to build foundation models to test their capabilities in broader applications, and examples include Prithvi by NASA-IBM, Segment-Anything-Model, ViT, etc.
This leads to an important question: 
Are foundation models always a suitable choice for different remote sensing tasks, and when or when not?
This work aims to enhance the understanding of the status and suitability of foundation models for pixel-level classification using multispectral imagery at moderate resolution, through comparisons with traditional machine learning (ML) and regular-size deep learning models.
Interestingly, the results reveal that in many scenarios traditional ML models still have similar or better performance compared to foundation models, especially for tasks where texture is less useful for classification. 
On the other hand, deep learning models did show more promising results for tasks where labels partially depend on texture (e.g., burn scar), while the difference in performance between foundation models and deep learning models is not obvious.
The results conform with our analysis: The suitability of foundation models depend on the alignment between the self-supervised learning tasks and the real downstream tasks, and the typical masked autoencoder paradigm is not necessarily suitable for many remote sensing problems.
\end{abstract}


\section{Introduction}
Foundation models are very large deep learning models that aim to generalize better with hundreds of millions, billions or even trillions of parameters.
These large models have recently demonstrated surprisingly well performances in various language and vision tasks, with strong generalization abilities across tasks.
In particular, GPT-type of models showed revolutionary capabilities in generative tasks and started to transform numerous aspects of existing routines in education, research, government and industry.
The unprecedented impacts brought by the foundation models naturally raise high expectations of their performances in other domains and motivate the developments of prototypes and domain-specific customizations.

Traditionally, the training of foundation models is extremely challenging, as the potential of the super-sized models comes together with the demand for super-sized training datasets, which is largely unavailable at the needed scales. 
Fortunately, the developments of self-supervised learning (SSL) came into rescue for various types of tasks such as generation. SSL tackles the data challenge for a real downstream task by creating related "artificial" tasks where the training data -- both the input features $\vx$ and artificial targets $\vy$ -- 
can be automatically generated at a large scale without tedious and time-consuming manual efforts. 
For example, masked modeling
is one of the most common SSL tasks, where parts of sentences or images are masked (i.e., left empty) from the original complete ones to be used as prediction targets for the SSL task \cite{he2022masked}. Because such target labels can be automatically generated in huge quantities, they can be used to pre-train the gigantic amount of parameters in foundation models. The pre-trained models can then be finetuned, or used with prompt-based strategies, for downstream tasks.

This paper aims to make a comparison between foundation models, regular-size deep learning models, and traditional machine learning (ML) models to understand the status and suitability of these models for pixel-level classification tasks using multispectral imagery at moderate spatial resolution (e.g., 10-30 m). Such tasks are both very common and highly important for critical sectors including agriculture, water, natural disaster management, etc.
Several efforts have been made recently to enhance the capabilities of foundation models for Earth science problems.
For example, prior studies such as FourCastNet \cite{pathak2022fourcastnet} and GraphCast \cite{lam2023learning} have shown the potential to accelerate weather forecasting models by hundreds of times while maintaining a very high accuracy. As most of these methods are designed to approximate physical Earth science models, they target physics-specific challenges (e.g., capturing global dependencies using the Fourier neural operator~\cite{li2020fourier}) but are not applicable to pixel-level classification tasks. 
Efforts have also been made to evaluate foundation models such as vision transformers for remote sensing data \cite{sun2022ringmo, jakubik2023toward, dias2023agenda}, and the larger models did show enhancements for scene classification and object detection on high-resolution aerial imagery. The most related open-source version is the foundation model Prithvi developed by NASA and IBM \cite{jakubik2023foundation, Prithvi-100M}, which is based on the multispectral Harmonized Landsat and Sentinel-2 (HLS) dataset.
However, pixel-level classification tasks using multispectral imagery at moderate spatial resolution exhibit differences from other typical vision tasks or remote sensing tasks on very high spatial resolution data. In particular, class labels may not bear a strong relationship with texture patterns in the images.
In this context, a simple yet important question that has not been addressed is whether foundation models such as Prithvi can offer significant improvements over traditional ML models such as random forest, which are still the models of choice for many pillar remote sensing products at national or global scales (e.g., \cite{glad, curtis2018classifying}).
More importantly, there is a lack of understanding regarding in what scenarios foundation models will have advantages over other learning models.

To better answer these questions, we consider three foundation models (including the most related Prithvi), three regular-size deep learning models, and four traditional ML models. 
Interestingly, the results reveal that in many scenarios traditional ML models still have a similar or better performance than foundation models. Particularly, they often work better for tasks (e.g., flood mapping) where textures are less useful for classification. On the other hand, deep learning models did show more promising results for tasks where labels partially depend on textures (e.g., burn scar), while the difference in performance between foundation models and regular-size models is not obvious.
These results conform with our analysis that 
the suitability of foundation models is subject to the alignment between the SSL tasks and the downstream tasks, and the typical masked autoencoder paradigm is not necessarily suitable for many remote sensing problems.

\section{Suitability of Foundation Models for Pixel-level Tasks}\label{sec:analysis}

While it is important to test out foundation model ideas (e.g., large-scale self-supervised pre-training) on remote sensing as a closely related application domain, it is equivalently important to enhance the understanding of the suitability of the models and their current status. This paper focuses on pixel-level classification tasks using multispectral imagery (e.g., Landsat-8, Sentinel-2), which are widely used in a wide range of applications such as agriculture monitoring, flood mapping, fire detection, land cover land use monitoring, etc.
In the following, we analyze the characteristics of foundation models and the implications of the designs for remote sensing tasks.

\subsection{Task Alignment: Pre-training vs. Downstream Tasks}
The alignment between the SSL tasks used in pre-training and the actual downstream tasks is an important factor in the performance of the finetuned foundation model \cite{xie2023geo, wei2021aligning}. 
Taking large language models (LLMs) as an example, their SSL tasks (a.k.a., pretext tasks) are often tightly aligned with the downstream tasks. Specifically, the SSL tasks utilize masked language modeling that masks out words or sentences as prediction targets (i.e., free labels), and let the network train on them to develop language generation abilities. Such SSL tasks are highly correlated to the actual downstream tasks, such as chatting, where the response is generated based on an input sentence (e.g., a question). 
Hence, the pre-training through these SSL tasks can often lead to substantial improvements for well-aligned downstream tasks. 

In vision problems, the masked autoencoding tasks \cite{he2022masked} that are typically used for self-supervised pre-training -- including remote sensing (e.g., Prithvi) -- are quite similar in the sense that they also try to mask out local image patches and let the network complete them.
However, the difference is the downstream task. While generating a masked-out sentence is very similar to generating a response, generating an image patch differs substantially from classifying the labels at the pixel-level (e.g., whether a pixel is a corn or soybean).
For high resolution images, this could still be very useful as the ability to generate image patches corresponds to the ability to generate and understand complex textures (e.g., the shape of a vehicle). This can help downstream tasks such as object detection or realistic image generation.
However, for many pixel-level classification tasks on multispectral imagery (e.g., Landsat-8), the textures at the resolution often do not carry as much useful information to distinguish different classes such as corn vs. soybean. In contrast, the signatures often come down to the spectral and temporal characteristics of the single pixels. This makes many pixel-level classification tasks similar to the classic tabular data problems. Interestingly, for tabular problems, traditional tree-based algorithms often still have better performance compared to deep learning models \cite{grinsztajn2022tree}.
This shows the importance of a comparison between 
foundation models and traditional ML approaches.
We hypothesize that traditional methods will be better suited when texture differences are less distinguishable between different classes, whereas foundation models may show advantages when textures do matter.

\subsection{Replicating Human Intelligence: Features vs. Labels}
The difference in the alignment between SSL and real downstream tasks in language and remote sensing problems is also related to another fundamental difference -- the data.
While both pre-training data for language problems (e.g., Wikipedia articles, books, code repositories) and remote sensing problems (e.g., satellite images) have huge volumes, there is a key difference when we take the goal of the models into consideration.
For both problems -- and most others in AI -- the goal is to imitate human intelligence, letting it be responding to our questions in ChatGPT or automatically classifying pixels as ``flooded" or ``not flooded."
With this in mind, we can see that the text data used for language problems serve the purpose well as they are in fact generated by humans, often with lots of thinking. Thus, when the SSL tasks mask out some sentences as prediction targets during pre-training, the labels (i.e., masked sentences) are human-generated labels that we want to imitate in downstream tasks such as text generation \cite{xie2023geo}.
This is fundamentally different for satellite data, where the masked-out image patches in the SSL tasks are not human annotations. In other words, the masked-out patches can only represent the input features in downstream tasks (e.g., input images for flood mapping) but not the actual human-generated labels that we need. %The bridge is missing.

\section{Foundation models}\label{sec:foundation}

This section briefly describes the architecture and training of the three foundation models used in the evaluation. The Vision Transformer (ViT) is a major backbone for foundation models \cite{dosovitskiy2020image,he2022masked}, which divides an image into fixed-size patches and order them in a sequence. Then it embeds these patches sequentially 
and feed the obtained embeddings into a standard transformer.
Such a learning process enables capturing long-range dependencies across the entire image. Following \cite{xie2021segformer}, a simplified deconvolutional segmentation head is used to replace the original classification head for pixel-level segmentation. ViT is pre-trained using masked autoencoding (MAE) as the SSL task on the ImageNet-21k.
The model is finetuned for each downstream task.

SegFormer is a variant of Transformer for semantic segmentation \cite{xie2021segformer}. Different from ViT, SegFormer consists of hierarchical Transformer encoders for capturing multi-scale features and a lightweight MLP decoder head for efficient segmentation to generate the pixel-level mask. SegFormer's encoder is pre-trained using an MAE on ImageNet-1k with 8 Tesla V100, and the segmentation head is trained on ADE20k. 
The model is further finetuned for each downstream task.

Prithvi is a ViT-based foundation model for remote sensing \cite{jakubik2023foundation, Prithvi-100M}.
It uses the ViT backbone with 3D positional encoding and MAE to better fit the multispectral bands and multi-temporal data. The lightweight decoder head consists of four deconvolutional layers and a convolutional layer.
Prithvi is pre-trained over 1TB of multispectral HLS dataset across the contiguous US, using up to 64 NVIDIA A100 GPUs, and then finetuned to each downstream task.

\section{Evaluation}

\subsection{Pixel-level classification tasks on multispectral imagery}
We use the same standard tasks and datasets as used in the evaluation of the NASA-IBM Prithvi \cite{jakubik2023foundation}, which is the most relevant foundation model for the scope of this paper 
as it is designed for multispectral imagery at moderate spatial resolution. 
The tasks include multi-temporal crop classification, wildfire scar mapping, and flood mapping. To be consistent, we adopt the same train-test splits used in \cite{jakubik2023foundation}. 
In \textbf{multi-temporal crop classification}\footnote{\url{https://huggingface.co/datasets/ibm-nasa-geospatial/multi-temporal-crop-classification}}, the imagery was collected from Sentinel satellites across the Contiguous United States in 2022, with the crop and land cover labels from USDA's Crop Data Layer. A total of 3,083 images were used for training and 771 images for testing, where each image contains 224$\times$224 pixels with 6 bands over 3 time steps. 
\textbf{Wildfire scar}\footnote{\url{https://huggingface.co/datasets/ibm-nasa-geospatial/hls_burn_scars}} data was gathered from HLS imagery aligned with the Monitoring Trends in Burn Severity historical fire database, where each HLS image was selected in the nearest date after one month of a fire event. The dataset included 804 HLS image chips of 512$\times$512 pixels and 6 spectral bands with a training-testing split of 67\%-33\%.
The dataset for \textbf{Flood mapping}\footnote{\url{https://github.com/cloudtostreet/Sen1Floods11}}, Sen1Floods11 \cite{bonafilia2020sen1floods11}, consists of raw Sentinel-1 and Sentinel-2 imagery covering 120,406 km\textsuperscript{2} and spanning all 14 biomes with 11 flood events at global scales. The dataset comprises 252 image chips for training, 89 for validating, and 90 for testing, each having 512$\times$512 pixels with 6 spectral bands.

\subsection{Experiment settings}
The following are the candidate methods used in the comparison among traditional ML models, regular-size deep learning models and foundation models. 
Both foundation and deep learning models take images as inputs, while traditional ML models use single-pixel inputs (with variants using local neighborhood augmentation).
\begin{itemize}[leftmargin=*]
    \itemsep0em
    \item \textbf{RF:} A random forest classifier with 100 trees.
    \item \textbf{RF$_{aug}$:} RF with augmented domain knowledge features including spectral indices (e.g., NDVI, NDWI) and spatial textures from a 3$\times$3 local neighborhood.
    \item \textbf{XGB:}
    The XGBoost
    classifier with 100 trees \cite{chen2016xgboost}.  
    \item \textbf{XGB$_{aug}$:} 
    XGB with augmented features (same as RF$_{aug}$).
    \item \textbf{FCN:} A fully-connected neural network with 5 layers.
    \item \textbf{U-Net:} An encoder-decoder model for segmentation \cite{ronneberger2015u}. 
    \item \textbf{DeepLabv3+ (D3+):} An encoder-decoder model with enhanced atrous convolution \cite{chen2018encoder}.
    \item \textbf{ViT:} The vision transformer foundation model in Sec. \ref{sec:foundation}.
    \item \textbf{SegFormer:} The segmentation-driven ViT in
    Sec. \ref{sec:foundation}.
    \item \textbf{Prithvi:} The foundation model designed for multispectral imagery at moderate resolution in Sec. \ref{sec:foundation}.
\end{itemize}

In the training stage, traditional ML models use default parameter settings from scikit-learn and XGBoost packages. Regular deep learning models were trained using the Adam optimizer for 100 epochs, and the pretrained foundational models were finetuned for 100 epochs. For Prithvi, we directly used the official finetuned models.
We evaluated the model performance using metrics including the Intesection-over-Union (IoU) and F1 score, which is the harmonic mean of precision and recall.
Instead of using a union of no-data pixels (areas outside the image boundaries) from the reference and the predicted results as masks for computing the metrics\cite{jakubik2023foundation}, we calculate the metrics in a more standard way 
using only no-data from the reference for masking, making sure the results will not be biased by incorrect no-data predictions (as a class).
In addition, clouds are originally considered as part of the no-data masks in flood and burn scar mapping. To provide a more comprehensive view, our results (Tables \ref{tab:burn} and \ref{tab:flood}) include two different versions with clouds as no-data masks, or, as a class.
In the tables, mIoU and mF1 are the average of the metrics computed using individual classes. 
% The overall accuracy, aAcc, is directly calculated over all the classes.

\begin{table*}[t]
\footnotesize
\caption{IoU Comparison for Multi-Temporal Crop Classification
\vspace{-6pt}
}
\begin{adjustbox}{width=1\textwidth}
\centering
\begin{tabular}{|l||ccccccccccccc|c|}
\hline
    Methods      & Nat. Veg. & Forest & Corn  & Soybeans & Wetlands & Developed & Water & Wheat & Alfalfa & Idle & Cotton & Sorghum & Other & mIoU  \\ 
                  \hline
RF                & 50.00                 & 53.51  & 58.53 & 55.42    &  \underline{43.84}    & 41.72            & \underline{76.48}      & 52.26        & 29.64   & 38.00                   & 35.22  & 38.44   & 38.85 & 47.07 \\
RF$_{aug}$      & \underline{50.25}              & \underline{53.76}  & \underline{59.56} & \underline{56.51}    & \textbf{44.20}     & \underline{42.00}               & 76.10       & \underline{53.24}        & 30.64   & \underline{38.68}                & 35.60   & 39.80    & \underline{39.72} & \underline{47.70}  \\
XGB           & 37.92              & 44.55  & 38.34 & 34.00       & 20.78    & 21.44            & 74.56      & 32.61        & 10.95   & 27.23                & 7.09   & 16.53   & 16.29 & 29.41 \\
XGB$_{aug}$ & 35.97              & 44.29  & 39.25 & 36.60     & 27.55    & 23.86            & 73.73      & 34.58        & 11.50    & 29.09                & 7.32   & 22.70    & 18.07 & 31.12 \\ \hline
FCN               & 37.82              & 42.22  & 33.53 & 29.39    & 21.94    & 14.04            & 73.84      & 29.47        & 7.76    & 21.34                & 0.16   & 13.06   & 8.02  & 25.58 \\
U-Net             & \textbf{55.21}      & \textbf{54.79}  & \textbf{64.29} & \textbf{63.08}    & 40.99    & \textbf{54.65}            & \textbf{76.86}      & \textbf{58.17}        & \textbf{39.12}   & \textbf{42.64}                & \textbf{42.26}  & \textbf{41.76}   & \textbf{42.20}  & \textbf{52.00}    \\
D3+        & 49.06              & 48.18  & 58.77 & 54.30     & 42.33    & 34.89            & 68.29      & 52.74        & \underline{36.07}   & 38.66                & \underline{39.55}  & \underline{40.18}   & 38.63 & 46.28 \\ 
\hline
ViT               & 44.39              & 41.63  & 50.07 & 49.22    & 37.69    & 28.33            & 68.34      & 41.44        & 24.74   & 32.22                & 29.47  & 32.68   & 29.78 & 39.23 \\
Segformer         & 47.14              & 45.10  & 52.95 & 51.30    & 37.14    & 32.05            & 69.34      & 46.69        & 29.00   & 34.58                & 31.86  & 35.20   & 33.73 & 42.01 \\
Prithvi     & 40.38              & 47.47  & 54.91 & 52.97    & 40.20     & 36.11            & 68.04      & 49.67        & 30.84   & 34.93                & 32.37  & 32.83   & 34.27 & 42.69 \\ 
\hline
\end{tabular}
\end{adjustbox}
\label{tab:crop_iou}
\vspace{-14pt}
\end{table*}

\begin{table*}[t]
\footnotesize
\caption{F1-score Comparison for Multi-Temporal Crop Classification
\vspace{-6pt}
}
\begin{adjustbox}{width=1\textwidth}
\centering
\begin{tabular}{|l||ccccccccccccc|c|}
\hline
     Methods      & Nat. Veg. & Forest & Corn  & Soybeans & Wetlands & Developed & Water & Wheat & Alfalfa & Idle & Cotton & Sorghum & Other & mF1 \\ \hline
RF                & 66.66              & 69.72  & 73.84 & 71.32    & \underline{60.95}    & 58.88            & \underline{86.67}      & 68.64        & 45.73   & 55.07                & 52.10   & 55.53   & 55.96 & 63.16 \\
RF$_{aug}$      & \underline{66.88}              & \underline{69.92}  & \underline{74.65} & \underline{72.21}    & \textbf{61.30}     & \underline{59.16}            & 86.42      & \underline{69.48}        & 46.91   & \underline{55.78}                & 52.51  & 56.94   & \underline{56.86} & \underline{63.77}\\
XGB           & 54.99              & 61.64  & 55.43 & 50.75    & 34.42    & 35.31            & 85.42      & 49.18        & 19.74   & 42.80                 & 13.24  & 28.37   & 28.02 & 43.02 \\
XGB$_{aug}$ & 52.91              & 61.39  & 56.37 & 52.59    & 43.20     & 38.53            & 84.88      & 51.39        & 20.62   & 45.07                & 13.63  & 37.01   & 30.61 & 45.32\\ \hline
FCN               & 54.88              & 59.37  & 50.22 & 45.43    & 35.99    & 24.62            & 84.95      & 45.52        & 14.41   & 35.17                & 0.32   & 23.11   & 14.85 & 37.60 \\
U-Net             & \textbf{71.14}              & \textbf{70.79}  & \textbf{78.27} & \textbf{77.36}    & 58.14    & \textbf{70.67}            & \textbf{86.92}      & \textbf{73.56}        & \textbf{56.24}   & \textbf{59.78}                & \textbf{59.41}  & \textbf{58.92}   & \textbf{59.36} & \textbf{67.74} \\
D3+        & 65.83              & 65.03  & 74.03 & 70.38    & 59.48    & 51.73            & 81.16      & 69.06        & \underline{53.01}   & 55.76                & \underline{56.68}  & \underline{57.33}   & 55.73 & 62.71 \\ 
\hline
ViT               & 61.49              & 58.79  & 66.73 & 65.97    & 54.75    & 44.15            & 81.19      & 58.59        & 39.67   & 48.73                & 45.52  & 49.26   & 45.90 & 55.44 \\
Segformer         & 64.08              & 62.16  & 69.24 & 67.81    & 54.16    & 48.54            & 81.89      & 63.66        & 44.96   & 51.39                & 48.32  & 52.07   & 50.45 & 58.36 \\
Prithvi     & 57.53              & 64.38  & 70.89 & 69.25    & 57.34    & 53.06            & 80.98      & 66.37        & 47.14   & 51.77                & 48.91  & 49.43   & 51.05 & 59.09 \\

\hline
\end{tabular}
\end{adjustbox}
\label{tab:crop_f1}
\vspace{-14pt}
\end{table*}

\begin{table}[t]
\footnotesize
\caption{Comparative Performance of All Models in Burn Scar
\vspace{-6pt}
}
\begin{adjustbox}{width=0.48\textwidth}
\centering
\setlength{\tabcolsep}{5pt} % Adjust column padding
\begin{tabular}{|l||cc|cc||cc|cc|}
\hline
& \multicolumn{4}{c||}{Clouds as no-data masks} &  \multicolumn{4}{c|}{Clouds as a class} \\\cline{2-9}%\hline
& \multicolumn{2}{c|}{Burn Scar} &  \multicolumn{2}{c||}{All classes} &
\multicolumn{2}{c|}{Burn Scar} & \multicolumn{2}{c|}{All classes} 
%  (w. cloud)
\\\hline
Methods     & IoU                                          & F1                                 & mIoU                                   & mF1                                & IoU                                       & F1                                 & mIoU                               & mF1                                \\ \hline
RF          & 58.86                                        & 74.1                               & 77.12                                  & 85.87                              & 58.85                                     & 74.1                               & 84.69                              & 90.55                              \\
RF$_{aug}$  & 64.21                                        & 78.2                               & 80.13                                  & 88.1                               & 64.2                                      & 78.19                              & 86.67                              & 92.02                              \\
XGB         & 48.98                                        & 65.76                              & 71.77                                  & 81.48                              & 48.95                                     & 65.73                              & 80.02                              & 87.06                              \\
XGB$_{aug}$ & 54.23                                        & 70.32                              & 74.63                                  & 83.89                              & 54.21                                     & 70.3                               & 81.85                              & 88.63                              \\ \hline
FCN         & 56.77                                        & 72.42                              & 76.01                                  & 84.99                              & 56.52                                     & 72.22                              & 81.07                              & 88.44                              \\
U-Net       & 76.95                                        & 86.97                              & 87.22                                  & 92.85                              & 76.87                                     & 86.92                              & 89.65                              & 94.29                              \\
D3+         & \textbf{81.34}             & \textbf{89.71}    & \textbf{89.6}         & \textbf{94.32}    & \textbf{81.24}           & \textbf{89.65}    & \textbf{91.3}     & \textbf{95.3}     \\ \hline
ViT         & \underline{78.77}           & \underline{88.12} & \underline{88.21}     & \underline{93.47} & \underline{78.64}        & \underline{88.04} & \underline{89.34} & \underline{94.18} \\
SegFormer   & 77.72                                        & 87.46                              & 87.59                                  & 93.09                              & 77.52                                     & 87.34                              & 88.86                              & 93.89                              \\
Prithvi     & 74.26                                        & 85.23                              & 85.53                                  & 91.8                               & 74.26                                     & 85.23                              & 86.6                               & 92.55                              \\

\hline
\end{tabular}
\end{adjustbox}
\label{tab:burn}
\vspace{-14pt}
\end{table}

\begin{table}[t]
\caption{Comparative Performance of All Models in Flood Mapping
\vspace{-6pt}
}
\begin{adjustbox}{width=0.48\textwidth}
\footnotesize
\centering
\setlength{\tabcolsep}{5pt} % Adjust column padding
\begin{tabular}{|l||cc|cc||cc|cc|}
\hline
& \multicolumn{4}{c||}{Clouds as no-data masks} &  \multicolumn{4}{c|}{Clouds as a class} \\\cline{2-9}%\hline
& \multicolumn{2}{c|}{Water} & 
\multicolumn{2}{c||}{All classes} & 
\multicolumn{2}{c|}{Water} & \multicolumn{2}{c|}{All classes}
% (w. cloud)
\\\hline
Methods     & IoU                                          & F1                                 & mIoU                                   & mF1                                & IoU                                   & F1                                 & mIoU                               & mF1                                \\ \hline
RF          & 79.70                                        & 88.7                               & 88.13                                  & 93.48                              & 78.42                                 & 87.9                               & 78.29                              & 87.32                              \\
RF$_{aug}$  & 81.00                                        & 89.5                               & 88.91                                  & 93.94                              & 79.74                                 & 88.73                              & \underline{79.79} & \underline{88.34} \\
XGB         & 82.02                                        & 90.12                              & 89.63                                  & 94.36                              & \underline{79.91}    & \underline{88.83} & 78.55                              & 87.49                              \\
XGB$_{aug}$ & \textbf{84.29}              & \textbf{91.47}    & \textbf{90.8}         & \textbf{95.05}    & \textbf{81.4}        & \textbf{89.75}    & \textbf{80.26}    & \textbf{88.63}    \\ \hline
FCN         & \underline{82.67}           & \underline{90.51} & \underline{89.68}     & \underline{94.42} & 79.6                                  & 88.64                              & 77.75                              & 86.91                              \\
U-Net       & 81.00                                        & 89.5                               & 87.18                                  & 93.03                              & 79.63                                 & 88.66                              & 77.72                              & 86.98                              \\
D3+         & 71.59                                        & 83.44                              & 83.65                                  & 90.63                              & 70.66                                 & 82.81                              & 57.03                              & 67.37                              \\ \hline
ViT         & 76.25                                        & 86.52                              & 86.23                                  & 92.23                              & 71.23                                 & 83.2                               & 70.21                              & 81.36                              \\
SegFormer   & 79.97                                        & 88.87                              & 86.25                                  & 92.50                              & 77.64                                 & 87.42                              & 76.12                              & 85.89                              \\
Prithvi     & 80.02                                        & 88.9                               & 88.5                                   & 93.69                              & 74.47                                 & 85.36                              & 58.59                              & 68.19                              \\
\hline

\end{tabular}
\end{adjustbox}
\label{tab:flood}
\vspace{-14pt}
\end{table}

\subsection{Results}

\subsubsection{Crop classification}
Tables \ref{tab:crop_iou} and \ref{tab:crop_f1} provide comparisons of IoU and F1 scores among foundation models, regular-size deep learning models, and traditional ML models for crop classification. 
Prithvi -- pre-trained with multispectral HLS data -- did have the best performance among the foundation models.
Though, interestingly, traditional ML models such as RF and RF$_{aug}$ outperformed it by about 4-5\%. 
The trend is similar across classes. 
The regular-size model U-Net had the best performance, indicating that the vast SSL pre-training from masked autoencoder in foundation models likely was not very helpful. As a result, the cons of the larger number of parameters (e.g., harder to train) may have largely outweighed the potential for better generalizability.
Overall, U-Net outperformed RF$_{aug}$, while RF$_{aug}$ outperformed Prithvi by about 4\%.
These 3 models, as summarized in Sec. \ref{sec:summary}, were the most stable within their model group (i.e., either the best or very close to the best in the group).
% , and overall accuracy

\subsubsection{Burn Scar}
Texture patterns and local contextual information are important for burn scar classification as ground exposed after burns could be confused with other normal ground if only pixel-level spectral patterns are considered. 
As a result, both foundation models and regular convolutional networks showed significant advantages over traditional models such as random forest (Table \ref{tab:burn}). DeepLabV3+ had the best results on this dataset, but the differences among deep learning models including foundation models (except the fully-connected FCN) are not significant. 
Additionally, DeepLabV3+ was less stable across the three datasets.
Traditional ML methods' lower performance shows their limitations in capturing texture patterns and contextual information beyond local pixels.
We also included the results where clouds are considered as a class instead of no-data masks in the evaluation.
This is because cloud pixels differ from other types of no-data (e.g., pixels falling outside the valid boundary of an image due to the non-orthogonal orientation of the image tile) and 
they may need to be predicted as part of the model outputs.
Here the trained models are the same as before, and the only difference is in the calculation of evaluation metrics. 
The results for single-class (water) also changed slightly because clouds are no longer used as no-data masks for computing metrics.
% when evaluating IoU and F1 scores.
For this burn scar task, the impact on ranking is less significant, and all scores increased (except FCN) potentially due to higher individual-class scores on clouds. However, the gaps between deep learning models and traditional ML models were reduced.

\subsubsection{Flood}
For flood classification, traditional ML methods showed the best performances (e.g., showing 2 out of the top 3 results for IoU and F1 scores; Table \ref{tab:flood}),
with XGB$_{aug}$ presenting the highest F1 and IoU among all models. This observation aligns with the fact that water exhibits a pronounced spectral difference compared to other ground types (e.g., can be well expressed by spectral indices such as NDWI \cite{thomas2023framework}).
Given that pixel-based spectral characteristics dominate -- while spatial textures are less pronounced -- for flood classification, a misalignment between the downstream task and the pre-training task becomes evident, consequently diminishing the effectiveness of the foundation model. In Table \ref{tab:flood}, the differences among the methods are not significant.  
However, the stability of the methods varied when evaluating the scenario with clouds included as a class.
Particularly, Prithvi, ViT, and DeepLabV3+ experienced relatively larger drops in their performance as the three methods significantly underestimated cloud pixels. 
% The other methods showed better stability by estimating a better distribution of the cloud class. 
The four traditional ML models took the top-4 mIoU and mF1 scores in this scenario.

\subsection{Summary across tasks}\label{sec:summary}
\begin{figure}%[!t]
\centering
\includegraphics[scale=0.38]{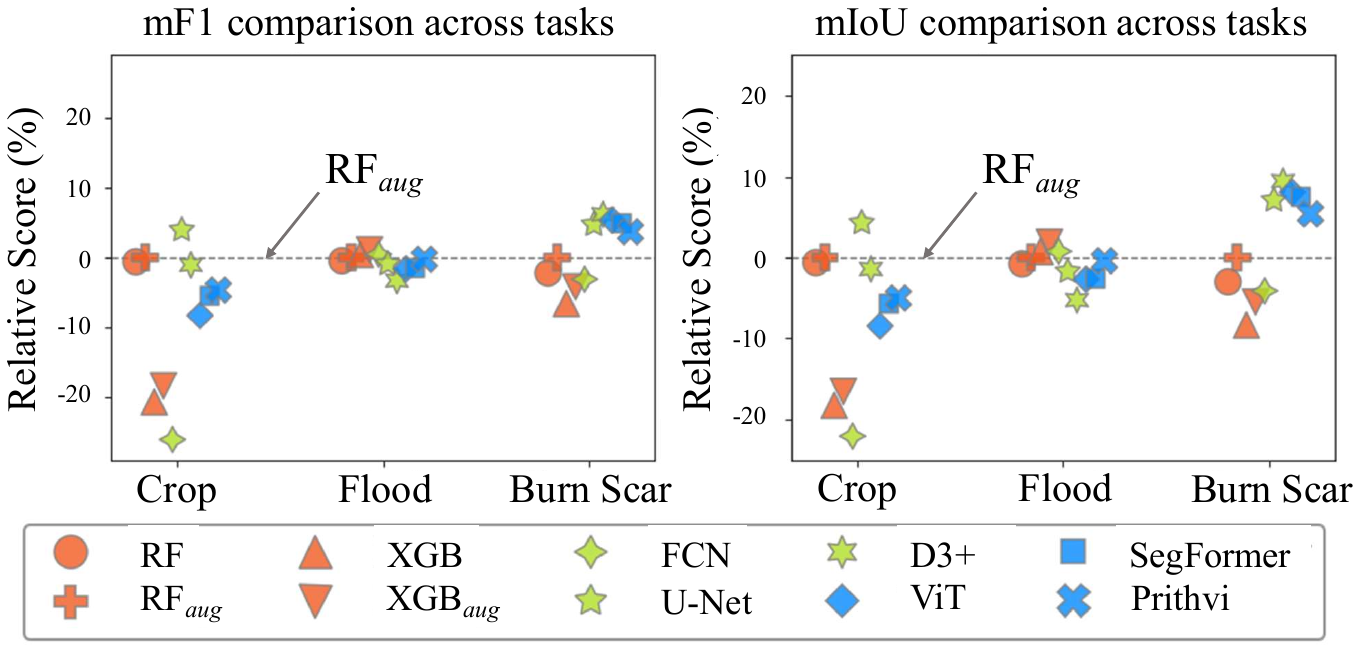}
\caption{Comparison across the 3 tasks using RF$_{aug}$ as a baseline. The methods are horizontally separated for better visual clarity.
}
\vspace{-7pt}
\label{fig:all}
\end{figure}

Fig. \ref{fig:all} shows the comparison across the three tasks, where the performance of RF$_{aug}$ is used as a baseline. For flood and burn scar, the results shown were the versions with clouds as no-data masks. Among the three groups of models, RF$_{aug}$, U-Net, and Prithvi, showed the most stable performances, respectively. They were either the best models in each group or within about 1-2\% compared to the best model. In contrast, other models experienced large variations in their performance. Other than burn scar classification where texture patterns are important, foundation models have not outperformed traditional models that are still widely used in major products (e.g., \cite{glad, curtis2018classifying}). When texture patterns are not as important, FCN (each pixel as an input) did not outperform traditional methods. Overall, we found U-Net to have the best performance, which likely strikes a good balance between model size and robustness against local noisy fluctuations (e.g., salt and pepper errors in traditional methods).
The comparison indicated that it is still premature to expect a meaningful paradigm shift by foundation models for pixel-level classification using multispectral imagery at moderate spatial resolution.

% \vspace{-7pt}
\section{Conclusion}
This paper aims to analyze the status and suitability of foundation models for pixel-level classification using multispectral imagery at moderate spatial resolution. 
The results conformed to our analysis: the standard masked autoencoder methods used for self-supervised pre-training only showed suitability when class labels have a high dependency on texture patterns from the images. As a result, 
the foundation models still did not outperform (and underperformed) traditional ML models on tasks where textures are not as informative.
However, foundation models showed effectiveness where texture features are important, but did not show improvements over regular-size deep learning models. This could be caused by the less complex and pronounced texture signatures, compared to those in typical vision problems.
Thus far, the results have not indicated a paradigm shift. However, efforts such as Prithvi are important to facilitate further understanding. Future efforts are needed for SSL tasks that are more aligned with real downstream tasks.
% From that perspective, 

\section*{Acknowledgments}
This material is based upon work supported by the National Science Foundation under Grant No. 2105133, 2126474, 2147195, 2118329, 1942680, 1952085, and 2021871; NASA under Grant No. 80NSSC22K1164 and 80NSSC24K0354; USGS under Grant No. G21AC10207; Google's AI for Social Good; the Zaratan supercomputing cluster at the University of Maryland; and Pitt Momentum Funds award and CRC at the University of Pittsburgh.

\bibliographystyle{IEEEtran}
\bibliography{ref}

\vfill

\end{document}